\documentclass[10pt, a4paper]{article}
\usepackage{lrec2016}
\usepackage{multibib}
\newcites{languageresource}{Language Resources}
\usepackage{graphicx}

 \usepackage{multirow}
 \usepackage[normalem]{ulem}
 \useunder{\uline}{\ul}{}
 
\usepackage{epstopdf}
\usepackage[latin1]{inputenc}

\usepackage{array}
\usepackage{algorithmic}
\usepackage{authblk}
\PassOptionsToPackage{hyphens}{url}
\usepackage{hyperref}
\usepackage{balance}

\newcommand{\thickhline}{%
    \noalign {\ifnum 0=`}\fi \hrule height 1pt
    \futurelet \reserved@a \@xhline
}

\title{Twitter as a Lifeline: Human-annotated Twitter Corpora for NLP of Crisis-related Messages}

\name{Muhammad Imran\textsuperscript{1}, Prasenjit Mitra\textsuperscript{1}, Carlos Castillo\textsuperscript{2}}

\address{~\textsuperscript{1}Qatar Computing Research Institute (HBKU), Doha, Qatar\\
\textsuperscript{2}Eurecat, Barcelona, Spain\\
mimran@qf.org.qa, pmitra@qf.org.qa, chato@acm.org\\}

\abstract{Microblogging platforms such as Twitter provide active communication channels during mass convergence and emergency events such as earthquakes, typhoons. During the sudden onset of a crisis situation, affected people post useful information on Twitter that can be used for situational awareness and other humanitarian disaster response efforts, if processed timely and effectively. Processing social media information pose multiple challenges such as parsing noisy, brief and informal messages, learning information categories from the incoming stream of messages and classifying them into different classes among others. One of the basic necessities of many of these tasks is the availability of data, in particular human-annotated data. In this paper, we present human-annotated Twitter corpora collected during 19 different crises that took place between 2013 and 2015. To demonstrate the utility of the annotations, we train machine learning classifiers. Moreover, we publish first largest word2vec word embeddings trained on 52 million crisis-related tweets. To deal with tweets language issues, we present human-annotated normalized lexical resources for different lexical variations.\\
~\\
\Keywords{Natural language processing, Twitter, Disaster response, Supervised classification} }

\begin{document} 
\maketitleabstract

\section{Introduction}
Twitter has been extensively used as an active communication channel, especially during mass convergence events such as natural disasters like earthquakes, floods, typhoons~\cite{imran2015processing,hughes2009twitter}. During the onset of a crisis, a variety of information is posted in real-time by affected people; by people who are in need of help (e.g., food, shelter, medical assistance, etc.) or by people who are willing to donate or offer volunteering services. Moreover, humanitarian and formal crisis response organizations such as government agencies, public health care NGOs, and military are tasked with responsibilities to save lives, reach people who need help, etc.~\cite{vieweg2014integrating}. Situation-sensitive requirements arise during such events and formal disaster response agencies look for actionable and tactical information in real-time to effectively estimate early damage assessment, and to launch relief efforts accordingly. 

Recent studies have shown the importance of social media messages to enhance situational awareness and also indicate that these messages contain significant actionable and tactical information~\cite{cameron2012emergency,imran2013extracting,purohit2013emergency}. Many Natural-Language-Processing (NLP) techniques such as automatic summarization, information classification, named-entity recognition, information extraction can be used to process such social media messages~\cite{bontcheva2013twitie,imran2015processing}. 
However, many social media messages are very brief, informal, and often contain slangs, typograpical errors, abbreviations, and incorrect grammar~\cite{han2013lexical}. These issues degrade the performance of many NLP techniques when used down the processing pipeline~\cite{ritter2010unsupervised,foster2011hardtoparse}. 

We present Twitter corpora consisting of more than 52 million crisis-related messages collected during 19 different crises. 
We provide human annotations (volunteers and crowd-sourced workers) of two types.  First,
the tweets are annotated with a set of categories such as displaced people, financial needs, infrastructure, etc.
These annotation schemes were built using input taken from formal crisis response agencies such as United Nations Office for the Coordination of Humanitarian Affairs (UN OCHA). 
Second, the tweets are annotated to identify out-of-vocabulary(OOV) terms, such as slangs, places names, abbreviations, misspellings, etc.
and their corrections and normalized forms.
This dataset can form the basis for research in text classification for short messages and for research on normalizing informal language.

Creating large corpora for training supervised machine-learning models is hard because it requires time and money that may not be available.
However, since our dataset was used for disaster relief efforts, volunteers were willing to annotate it; this work can now be leveraged to improve text classification and language processing tasks. 
Our work provides annotations for around 50,000 thousand messages, which is a significant corpus, that will enable research into applied machine learning and consequently benefit the disaster relief (and other) research communities.
Our dataset has been collected from various countries and during various times of the year.  This diversity would make it an interesting dataset that if used would be a foil to solutions that only work for specific language ``dialects", e.g., American English and would fail or suffer from degradation of quality if applied to variations, such as Indian English.
Our work shows that when a dataset is used for a real application, we could obtain larger number of annotations than otherwise.
These can then be used to improve text processing as a by-product.

The annotated data is also used to train machine-learning classifiers. In this case, we use three well-known learning algorithms: Naive Bayes, Random Forest, and Support Vector Machines (SVM). 
We remark that these classifiers are useful for formal crisis response organizations as well as for the research community to build more effective computational methods~\cite{pak2010twitter,imran2015processing} on top. We also train word2vec word embeddings from all 52 million messages and make them available to research community.

\subsection{Contributions}
The contributions of this paper are as follows:
\begin{enumerate}
\item We present human-annotated crisis-related messages collected during 19 different crises
\item We use human-annotations to built machine-learning classifiers in a multiclass classification setting to classify messages that are useful for humanitarian efforts
\item We provide first largest word2vec word embeddings trained using 52 million crisis-related messages

\item We use the collected data to identify OOV (out-of-vocabulary) words and provide human-annotated normalized lexical resources for different lexical variations
\end{enumerate}

\subsection{Paper organization}
The rest of the paper is organized as follows. In the next section, we describe datasets details and annotation schemes. Section 3 describes supervised classification task and word2vec word embeddings. Section 4 provides details of text normalization and we present related work in section 5. We conclude the paper in section 6.
\section{Crises Corpora Collection and Annotation}
\subsection{Data collection}
We collected crisis-related messages from Twitter posted during 19 different crises that took place from 2013 to 2015. Table~\ref{dataset_tbl} shows the list of crisis events along with their names, crisis type (e.g. earthquake, flood), countries where they took place, and the number of tweets each crisis contains. We collected these messages using our AIDR (Artificial Intelligence for Disaster Response) platform~\cite{imran2014aidr}. AIDR is an open source platform to collect and classify Twitter messages during the onset of a humanitarian crisis. AIDR has been used by UN OCHA during many major disasters such as Nepal Earthquake, Typhoon Hagupit. 

AIDR provides different convenient ways to collect messages from Twitter using the Twitter's streaming API. One can use different data collection strategies. For example, collecting tweets that contain some keywords and are specifically from a particular geographical area/region/city (e.g. New York). The detailed data collection strategies used to collect the datasets shown in Table~\ref{dataset_tbl} are included in each dataset folder.

\begin{table*}[ht!]
\small
\centering
\caption{Crises datasets details including crisis type, name, year, language of messages, country, \# of tweets.}
\label{dataset_tbl}
\begin{tabular}{llllrrrl}
\hline
{\bf Crisis type} & {\bf Crisis name}                & {\bf Country}  & {\bf Language} & \multicolumn{1}{l}{{\bf \# of Tweets}} & {\bf Start-date} & {\bf End-date} \\ \hline 
Earthquake        & Nepal Earthquake                 & Nepal                                   & English        & 4,223,937                              & 2015-04-25       & 2015-05-19     \\
Earthquake        & Terremoto Chile                  & Chile                                    & Spanish        & 842,209                                & 2014-04-02       & 2014-04-10     \\
Earthquake        & Chile Earthquake                 & Chile                                    & English        & 368,630                                & 2014-04-02       & 2014-04-17     \\
Earthquake        & California Earthquake            & USA                                      & English        & 254,525                                & 2014-08-24       & 2014-08-30     \\
Earthquake  &   Pakistan Earthquake & Pakistan   &English   &   156,905 &  2013-09-25 & 2013-10-10 \\
Typhoon           & Cyclone PAM                      & Vanuatu                                  & English       & 490,402                                & 2015-03-11       & 2015-03-29     \\
Typhoon           & Typhoon Hagupit                  & Phillippines                             & English        & 625,976                                & 2014-12-03       & 2014-12-16     \\
Typhoon           & Hurricane Odile        & Mexico                                   & English        & 62,058                                 & 2014-09-15       & 2014-09-28     \\
Volcano           & Iceland Volcano                  & Iceland                                  & English        & 83,470                                 & 2014-08-25       & 2014-09-01  \\     
Landslide         & Landslides worldwide        & Worldwide                             & English        & 382,626                                & 2014-03-12       & 2015-05-28     \\
Landslide         & Landslides worldwide        & Worldwide                                & French         & 17,329                                 & 2015-03-12       & 2015-06-23     \\
Landslide         & Landslides worldwide       & Worldwide                                & Spanish        & 75,244                                 & 2015-03-12       & 2015-06-23     \\
Floods  & Pakistan Floods   &   Pakistan        &   English &   1,236,610   &   2014-09-07 & 2014-09-22  \\ 
Floods  & India Floods   &   India        &   English &   5,259,681  &   2014-08-10 & 2014-09-03 \\
War \& conflict  &   Palestine Conflict  &   Palestine    &   English &   27,770,276  &   2014-07-12  &   2014-10-02 \\ 
War \& conflict & Peshawar Attack Pakistan & Pakistan  & English & 1,135,655 & 2014-12-16 & 2014-12-28 \\
Biological        & Middle East Respiratory Syndrome & Worldwide                                & English       & 215,370                                & 2014-04-27       & 2014-07-14     \\
Infectious disease  &   Ebola virus outbreak &   Worldwide  & English &   5,107,139 &  2014-08-02 & 2014-10-27 \\
Airline accident    & Malaysia Airlines flight MH370 & Malaysia                                 & English       & 4,507,157                              & 2014-03-11       & 2014-07-12     \\
\hline  
\end{tabular}
\end{table*}

\subsection{Data annotation}
Messages posted on social media vary greatly in terms of information they contain. 
For example, users post messages of personal nature, messages useful for situational awareness (e.g. infrastructure damage, causalities, individual needs), or not related to the crisis at all. 
Depending on their information needs, different humanitarian organizations use different annotation schemes to categories these messages. In this work, we use a subset of the annotations used by the United Nations Office for the Coordination of Humanitarian Affairs (UN OCHA). The 9 category types (including two catch-all classes: ``Other Useful Information" and ``Irrelevant") used by the UN OCHA are shown in the below-presented annotation scheme. For most of the datasets we have performed annotations by employing volunteers and paid workers.

To perform volunteered-based annotations, messages were collected from Twitter in real-time and passed through a de-duplication process. Only unique messages were considered for human-annotation. We use Stand-By-Task-Force (SBTF)\footnote{\url{http://blog.standbytaskforce.com/}} volunteers to annotate 
messages using our MicroMappers platform.\footnote{\url{http://micromappers.org/}} The real-time annotation process helps train machine learning classifiers rapidly, which are then used to classify new incoming messages. This process helps address time-critical information needs requirement of many humanitarian organizations.

After the first round of annotations, we found that some categories are small in terms of number of labels thus showing high class-imbalance. A dataset is said to be {\it imbalanced} if at least one of the classes has significantly fewer annotated instances than the others. The class imbalance problem has been known to hinder the learning performance of classification algorithms. In this case, we performed another round of annotations for datasets that have high class imbalance using the paid crowdsourcing platform CrowdFlower.\footnote{\url{http://crowdflower.com/}} 

In both annotation processes, an annotation task consists of a tweet and the list of categories listed below. A paid worker or volunteer reads the message and selects one of the categories most suitable for the message. Messages that do not belong to any category but contain some important information are categorized as ``Other Useful Information". A task is finalized (i.e. a category is assigned) when three different volunteers/paid workers agree on a category. 

According to the Twitter's data distribution policy, we are not allowed to publish actual contents of more than 50k tweets. For this reason, we publish all annotated tweets, which are less than 50k, along with tweet-ids of all the unannotated messages at {\url{http://CrisisNLP.qcri.org/}}. We also provide a tweets retrieval tool implemented in Java, which can be used to get full tweets content from Twitter. 

In below we show the annotation scheme used for crisis events caused by natural disasters. For other events, details regarding their annotations are available with the published data.\smallskip

{\bf Annotation scheme: Categorizing messages by information types}

\begin{itemize}
\item {\it Injured or dead people}: Reports of casualties and/or injured people due to the crisis
\item {\it Missing, trapped, or found people}: Reports and/or questions about missing or found people
\item {\it Displaced people and evacuations}: People who have relocated due to the crisis, even for a short time (includes evacuations)
\item{\it Infrastructure and utilities damage}: Reports of damaged buildings, roads, bridges, or utilities/services interrupted or restored
\item {\it Donation needs or offers or volunteering services}: Reports of urgent needs or donations of shelter and/or supplies such as food, water, clothing, money, medical supplies or blood; and volunteering services
								
\item {\it Caution and advice}: Reports of warnings issued or lifted, guidance and tips
					
\item{\it Sympathy and emotional support}: Prayers, thoughts, and emotional support
					
\item{\it  Other useful information}: Other useful information that helps understand the situation
					
\item{\it Not related or irrelevant}: Unrelated to the situation or irrelevant 

\end{itemize}




\section{Classification of Messages}
To make sense of huge amounts of Twitter messages posted during crises, we consider a basic operation, 
that is, the automatic categorization of messages into the categories of interest. This is a multiclass categorization problem in which instances are categorized into one of several classes. Specifically, we aim at learning a predictor $h : \mathcal{X} \rightarrow \mathcal{Y}$, where $\mathcal{X}$ is the set of messages and $\mathcal{Y}$ is a finite set of categories. For this purpose, we use three well-known learning algorithms i.e. Naive Bayes (NB), Support Vector Machines (SVM), and Random Forest (RF).

\subsection{Preprocessing and feature extraction} 
Prior to learning a classifier, we perform the following preprocessing steps. First, stop-words, URLs, and user-mentions are removed from the Twitter messages. We perform stemming using the Lovins stemmer. We use Uni-grams and bi-grams as our features. Previous studies found these two features outperform when used for similar classification tasks~\cite{imran2013extracting}. Finally, we used the information gain, a well-know feature selection method to select top 1k features. The labeled data we used in this task was annotated by the paid workers.

\subsection{Evaluation and Results}
We trained all three different kinds of classifiers using the preprocessed data. For the evaluation of the trained models, we used 10-folds cross-validation technique. Table~\ref{classification_res} shows the results of the classification task in terms of Area Under ROC curve\footnote{\url{https://en.wikipedia.org/wiki/Receiver_operating_characteristic}} for all classes of the 8 different disaster datasets. We also show the proportion of each class in each dataset.

Given the complexity of the task i.e. multiclass classification of short messages, we can see that all three classifiers have pretty decedent results. In this case, a random classifier represents an AUC = 0.50 and higher values are preferable. Other than the ``missing trapped or found people" class, which is the smallest class in term of proportion across all the datasets, results for most of the other classes are at the acceptable level (i.e. $\geq$ 0.80).

\begin{table*}[]
\centering
\scriptsize
\caption{Classification results in terms of Area Under ROC Curve for selected datasets across all classes using Support Vector Machines (SVM), Naive Bayes (NB), and Random Forest (RF).}
\label{classification_res}
\begin{tabular}{llrrrrrrrrr}
\hline
\multicolumn{1}{l|}{Datasets}        & \multicolumn{1}{l|}{Classifier} & \multicolumn{1}{l|}{\begin{tabular}[c]{@{}l@{}}Caution \\ and \\ advice\end{tabular}} & \multicolumn{1}{l|}{\begin{tabular}[c]{@{}l@{}}Displaced \\ people and \\ evacuations\end{tabular}} & \multicolumn{1}{l|}{\begin{tabular}[c]{@{}l@{}}Donation \\ needs or \\ offers\end{tabular}} & \multicolumn{1}{l|}{\begin{tabular}[c]{@{}l@{}}Infrastructure \\ and utilities \\ damage\end{tabular}} & \multicolumn{1}{l|}{\begin{tabular}[c]{@{}l@{}}Injured or \\ dead people\end{tabular}} & \multicolumn{1}{l|}{\begin{tabular}[c]{@{}l@{}}Missing trapped \\ or found people\end{tabular}} & \multicolumn{1}{l|}{\begin{tabular}[c]{@{}l@{}}Sympathy \\ emotional \\ support\end{tabular}} & \multicolumn{1}{l|}{\begin{tabular}[c]{@{}l@{}}Other useful \\ information\end{tabular}} & \multicolumn{1}{l}{\begin{tabular}[c]{@{}l@{}}Not related \\ or irrelevant\end{tabular}} \\ \hline \hline
\multirow{4}{*}{\begin{tabular}[c]{@{}l@{}}2014 Chile\\ earthquake\end{tabular}}        & Size(\%)                     & 15\%                                                                               & 2.80\%                                                                                              & 0.76\%                                                                                      & 1.70\%                                                                                                 & 5.60\%                                                                                 & 0.54\%                                                                                          & 25\%                                                                                           & 30\%                                                                                     & 19\%                                                                                      \\ 
                                      & SVM                          & 0.87                                                                               & 0.89                                                                                                & 0.57                                                                                        & 0.90                                                                                                    & 0.97                                                                                   & 0.23                                                                                            & 0.93                                                                                           & 0.86                                                                                     & 0.93                                                                                      \\ 
                                      & NB                           & 0.86                                                                               & 0.93                                                                                                & 0.78                                                                                        & 0.88                                                                                                   & 0.97                                                                                   & 0.64                                                                                            & 0.93                                                                                           & 0.87                                                                                     & 0.95                                                                                      \\
                                      & RF                           & 0.83                                                                               & 0.86                                                                                                & 0.67                                                                                        & 0.74                                                                                                   & 0.96                                                                                   & 0.46                                                                                            & 0.94                                                                                           & 0.86                                                                                     & 0.92                                                                                      \\ \hline
\multirow{4}{*}{\begin{tabular}[c]{@{}l@{}}2015 Nepal\\ earthquake\end{tabular}}         & Size(\%)                     & 2.10\%                                                                             & 3.10\%                                                                                              & 28\%                                                                                        & 4.50\%                                                                                                 & 11\%                                                                                   & 5.80\%                                                                                          & 17\%                                                                                           & 22\%                                                                                     & 6.50\%                                                                                    \\
                                      & SVM                          & 0.47                                                                               & 0.80                                                                                                 & 0.89                                                                                        & 0.85                                                                                                   & 0.95                                                                                   & 0.86                                                                                            & 0.88                                                                                           & 0.76                                                                                     & 0.75                                                                                      \\
                                      & NB                           & 0.68                                                                               & 0.82                                                                                                & 0.91                                                                                        & 0.90                                                                                                    & 0.95                                                                                   & 0.89                                                                                            & 0.91                                                                                           & 0.79                                                                                     & 0.84                                                                                      \\
                                      & RF                           & 0.56                                                                               & 0.73                                                                                                & 0.89                                                                                        & 0.74                                                                                                   & 0.94                                                                                   & 0.87                                                                                            & 0.89                                                                                           & 0.76                                                                                     & 0.75                                                                                      \\ \hline
\multirow{4}{*}{\begin{tabular}[c]{@{}l@{}}2013 Pakistan\\ earthquake\end{tabular}}      & Size(\%)                     & 6.30\%                                                                             & 0.82\%                                                                                              & 15\%                                                                                        & 2\%                                                                                                    & 17\%                                                                                   & 0.49\%                                                                                          & 5.60\%                                                                                         & 35\%                                                                                     & 18\%                                                                                      \\
                                      & SVM                          & 0.77                                                                               & 0.80                                                                                                & 0.92                                                                                        & 0.76                                                                                                   & 0.95                                                                                   & 0.63                                                                                            & 0.82                                                                                           & 0.84                                                                                     & 0.84                                                                                      \\
                                      & NB                           & 0.82                                                                               & 0.87                                                                                                & 0.94                                                                                        & 0.91                                                                                                   & 0.93                                                                                   & 0.74                                                                                            & 0.83                                                                                           & 0.84                                                                                     & 0.84                                                                                      \\
                                      & RF                           & 0.68                                                                               & 0.70                                                                                                & 0.92                                                                                        & 0.77                                                                                                   & 0.95                                                                                   & 0.69                                                                                            & 0.78                                                                                           & 0.88                                                                                     & 0.83                                                                                      \\ \hline
\multirow{4}{*}{\begin{tabular}[c]{@{}l@{}}2015 Cyclone\\ Pam\end{tabular}}      & Size(\%)                     & 7\%                                                                                & 3.10\%                                                                                              & 17\%                                                                                        & 11\%                                                                                                   & 7.20\%                                                                                 & 1.30\%                                                                                          & 5\%                                                                                            & 25\%                                                                                     & 24\%                                                                                      \\
                                      & SVM                          & 0.76                                                                               & 0.80                                                                                                 & 0.92                                                                                        & 0.85                                                                                                   & 0.95                                                                                   & 0.39                                                                                            & 0.66                                                                                           & 0.77                                                                                     & 0.90                                                                                       \\
                                      & NB                           & 0.79                                                                               & 0.82                                                                                                & 0.92                                                                                        & 0.86                                                                                                   & 0.97                                                                                   & 0.56                                                                                            & 0.79                                                                                           & 0.80                                                                                      & 0.94                                                                                      \\
                                      & RF                           & 0.68                                                                               & 0.80                                                                                                 & 0.90                                                                                         & 0.80                                                                                                    & 0.95                                                                                   & 0.47                                                                                            & 0.71                                                                                           & 0.79                                                                                     & 0.92                                                                                      \\ \hline
\multirow{4}{*}{\begin{tabular}[c]{@{}l@{}}2014 Typhoon \\ Hagupit\end{tabular}} & Size(\%)                     & 20\%                                                                               & 6.60\%                                                                                              & 5.50\%                                                                                      & 5.10\%                                                                                                 & 3\%                                                                                    & 0.58\%                                                                                          & 13\%                                                                                           & 33\%                                                                                     & 13\%                                                                                      \\
                                      & SVM                          & 0.74                                                                               & 0.95                                                                                                & 0.88                                                                                        & 0.76                                                                                                   & 0.94                                                                                   & 0.44                                                                                            & 0.92                                                                                           & 0.74                                                                                     & 0.81                                                                                      \\
                                      & NB                           & 0.75                                                                               & 0.96                                                                                                & 0.89                                                                                        & 0.82                                                                                                   & 0.96                                                                                   & 0.57                                                                                            & 0.92                                                                                           & 0.78                                                                                     & 0.81                                                                                      \\
                                      & RF                           & 0.71                                                                               & 0.97                                                                                                & 0.84                                                                                        & 0.73                                                                                                   & 0.94                                                                                   & 0.58                                                                                            & 0.91                                                                                           & 0.75                                                                                     & 0.80                                                                                       \\ \hline
\multirow{4}{*}{\begin{tabular}[c]{@{}l@{}}2014 India \\ floods\end{tabular}}     & Size(\%)                     & 3.60\%                                                                             & 1.40\%                                                                                              & 2.60\%                                                                                      & 4.30\%                                                                                                 & 47\%                                                                                   & 0.87\%                                                                                          & 1.30\%                                                                                         & 14\%                                                                                     & 25\%                                                                                      \\
                                      & SVM                          & 0.82                                                                               & 0.80                                                                                                 & 0.92                                                                                        & 0.92                                                                                                   & 0.97                                                                                   & 0.66                                                                                            & 0.63                                                                                           & 0.87                                                                                     & 0.97                                                                                      \\
                                      & NB                           & 0.89                                                                               & 0.92                                                                                                & 0.93                                                                                        & 0.90                                                                                                    & 0.93                                                                                   & 0.79                                                                                            & 0.83                                                                                           & 0.89                                                                                     & 0.98                                                                                      \\
                                      & RF                           & 0.83                                                                               & 0.79                                                                                                & 0.86                                                                                        & 0.87                                                                                                   & 0.97                                                                                   & 0.66                                                                                            & 0.65                                                                                           & 0.91                                                                                     & 0.96                                                               \\ \hline
                                      
\multirow{4}{*}{\begin{tabular}[c]{@{}l@{}}2014 Pakistan \\ floods\end{tabular}}     & Size(\%)                     
& 3.90\%		&	6.20\%	&	25\%	 &	5.40\%	&	13\%		&	6.40\%	&	6\%	&	32\% 	&	2.30\%
\\
& SVM           & 0.71	&	0.84 &	0.82 &	0.77 &	0.94	 & 0.85 &	0.88 &	0.74	 & 0.47 \\
& NB  & 0.83	 & 0.80 &	0.85	 & 0.79 &	0.94 &	0.85 &	0.89	 &0.77 & 	0.65  \\
& RF & 0.72 &	0.80 &	0.87 &	0.78 &	0.95 &	0.84 &	0.86 &	0.79 &	0.59
 \\ \hline
                                      
\multirow{4}{*}{\begin{tabular}[c]{@{}l@{}}2014 California \\ earthquake\end{tabular}}     & Size(\%)                     
& 6.30\% &	0.48\% & 	4.30\% & 	18\% &	10\% &	0.51\% &	 4.10\% &	47\% &	9.40\%
\\
& SVM & 0.84 &	0.54	 & 0.93 &	0.88	 & 0.97 &	0.62	 & 0.84 &	0.77 &	0.72\\
& NB  & 0.88 &	0.57	 & 0.94 &	0.86	 & 0.97 &	0.79	 & 0.90 &	0.78	 & 0.77 \\
& RF & 0.81 &	0.49 &	0.87 &	0.89 &	0.98 &	0.57	 & 0.88 &	0.81	 & 0.77
 \\ \hline \hline
\end{tabular}
\end{table*}

\subsection{Crisis word embeddings}
Many applications of machine learning and computational linguistics rely on semantic representations and relationships between words of a text document. Many different types of methods have been proposed that use continuous representations of words such as Latent Semantic Analysis (LSA) and Latent Dirichlet Allocation (LDA). However, recently models based on distributional representations of words become more famous. In this work, we train word embeddings (i.e. distributed word representations) using the 52 million Twitter messages in our datasets and make it available to research community. To the best of our knowledge this is the first largest word embeddings that are trained on crisis-related tweets.

We use word2vec, a very popular software to train word embedding~\cite{mikolov2013efficient}. As preprocessing, we replaced URLs, digits, and usernames with fixed constants and removed special characters. Finally, the word embeddings are generated using Continuous Bag Of Words (CBOW) architecture with negative sampling along with 300 word representation dimensionality. 

\section{Twitter Text Normalization}

\subsection{Language issues in Twitter messages}
The quality---in terms of readability, grammar, sentence structure etc.---of Twitter messages vary significantly. Typically, Twitter messages are brief, informal, noisy, unstructured, and often contain misspellings and grammatical mistakes. Moreover, due to Twitter's 140 character limit restriction, Twitter users intentionally shorten words by using abbreviations, acronyms, slangs, and sometimes words without spaces. The accuracy of natural language processing techniques would improve if we can identify the informal nature of the language in tweets and normalize OOV terms~\cite{han2013lexical}. We divide these lexical variations into the following five categories:
\begin{enumerate}
\item {\it Typos/misspellings:} e.g. earthquak (earthquake), missin (missing), ovrcme (overcome)
\item {\it Single-word abbreviation/slangs:} e.g. pls (please), srsly (seriously), govt (government), msg (message)
\item {\it Multi-word abbreviation/slangs:} e.g. imo (in my opinion), im (i am), brb (be right back)
\item {\it Phonetics substitutions:} e.g. 2morrow (tomorrow),  4ever (forever), 4g8 (forget), w8 (wait)
\item {\it Words without spaces:} e.g. prayfornepal (pray for nepal), wehelp (we help), weneedshelter (we need shelter)
\end{enumerate}

\begin{figure*}[htbp!]
\includegraphics[width=0.77\textwidth]{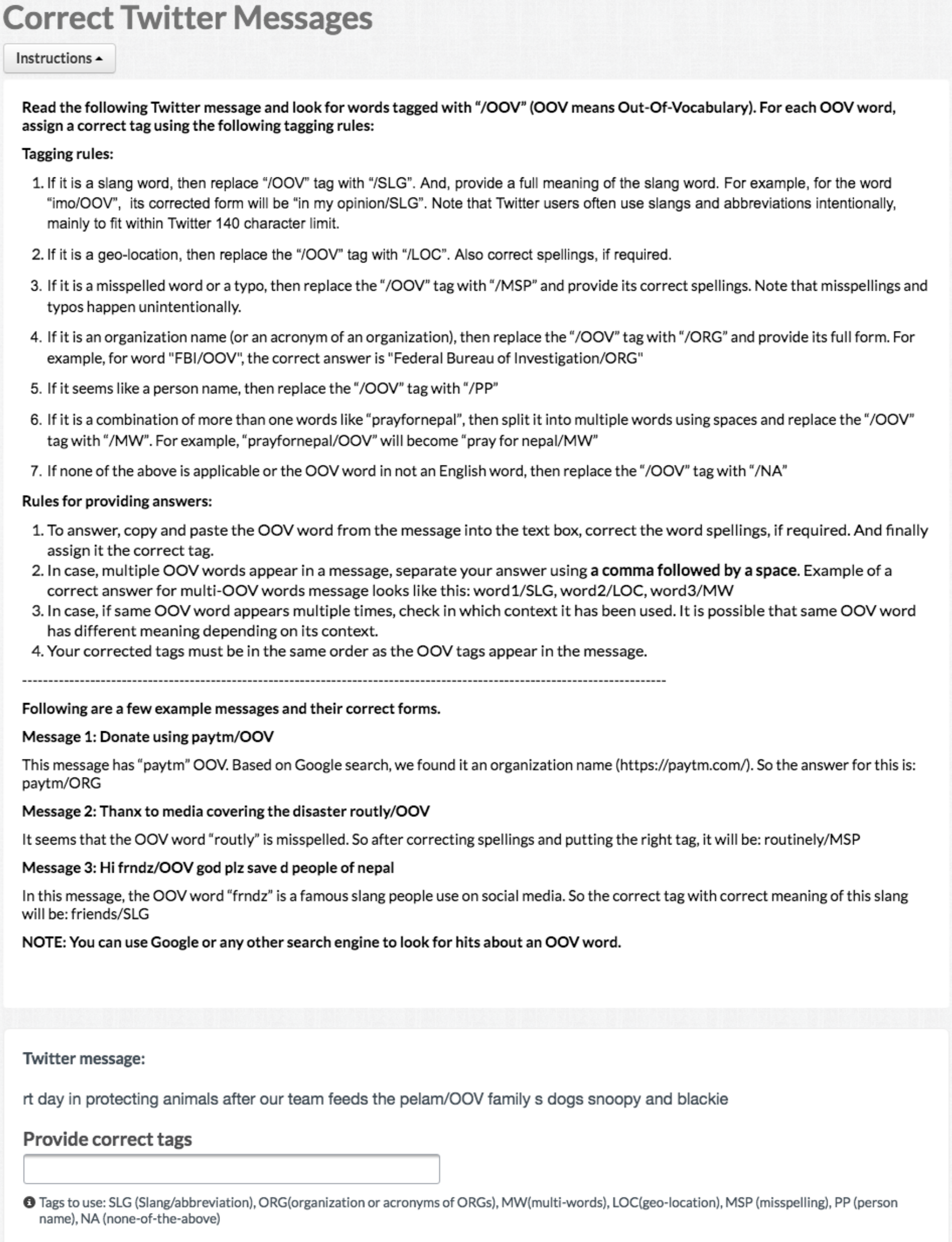}
\centering
\caption{Crowdsourcing task for Twitter out-of-vocabulary words normalization}
\label{fig:normaliz_task} 
\end{figure*}

\subsection{Identification of candidate OOV words}
To identify candidate OOV words that require normalization, we first build initial vocabularies consisting of lexical variations mentioned in the previous section. We use a dictionary available on the web to normalize abbreviations, chat shortcuts, and slang.\footnote{\url{http://www.innocentenglish.com/news/texting-abbreviations-collection-texting-slang.html}} 
We also use the SCOWL (Spell Checker Oriented Word Lists) aspell English dictionary~\footnote{\url{http://wordlist.aspell.net/}} that consists of 349,554 English words. 
The SCOWL dictionary is suitable for English spell checkers for most of English dialects. Although, the SCOWL dictionary contains places names (e.g. names of countries and famous cities),
after testing it on Nepal Earthquake data, we found that its coverage is not complete and a large number of cities/towns of Nepal are missing. To overcome this issue, we use 
the MaxMind~\footnote{\url{https://www.maxmind.com/en/free-world-cities-database}} world cities database that consists of 3,173,959 cities.

Using the above resources, we try to find OOV words in the dataset. However, we observed that a large number of OOVs consist of misspelled words for which a correct form can be obtained using one edit-distance change (i.e. by performing one insertion, deletion, or substitution operation). For this purpose, we train a language model using lists of most frequent words from Wiktionary,\footnote{\url{http://en.wiktionary.org/wiki/Wiktionary}} the British National Corpus,\footnote{\url{http://www.kilgarriff.co.uk/bnc-readme.html}} and words in our SCOWL dictionary. 
For a given misspelled word $w$, we aim to find a correction $c$ out of all possible corrections where the probability of $c$ given $w$ is maximum, i.e., $argmax_c P(c|w)$ \smallskip
By Bayes Theorem this is equivalent to:\smallskip

$argmax_c P(c|w) = argmax_c P(w|c) P(c)/P(w)$ \smallskip

or it can be written as: \smallskip

$argmax_c P(c|w) = argmax_c P(w|c) P(c)$ \smallskip

where $P(c)$ is the probability that $c$ is the correct word and $P(w|c)$ is the probability that the author typed $w$ when $c$ was intended. 
We then restrict the language model to predict corrections within one edit-distance range and from those choose the one with highest probability. 
Misspellings for which more than one change is required, we consider them as OOVs to be corrected by human workers.


\subsection{Normalization of OOV words}
To normalize the identified OOV words, we used the CrowdFlower crowdsourcing platform. A crowdsourcing task in this case consists of a Twitter message that contains one or more OOV words and a set of instructions shown in Figure~\ref{fig:normaliz_task}. The workers were asked to read the instructions and examples carefully before providing an answer. A worker reads the given message and provides a correct OOV tag (i.e. slang/abbreviation/acronym, a location name, an organization name, a misspelled word, or a person name). If an OOV is a misspelled word, the worker also provides its corrected form. We provide all the resources and the results of crowdsoucing to research community.




\section{Related Work}
The use of microblogging platforms such as Twitter during the sudden onset of a crisis situation has been increased in the last few years. Thousands of crisis-related messages that are posted online contain important information that can also be useful to humanitarian organizations for disaster response efforts, if processed timely and effectively~\cite{hughes2009twitter,imran2015processing}.

Many different types of processing techniques ranging from machine learning to natural language processing to computational linguistics have been developed~\cite{corvey2010twitter} for different purposes~\cite{imran2016enabling}. Despite there exists some resources e.g.~\cite{temnikova2015emterms,olteanu2015expect}, however, due to the scarcity of relevant data, in particular human-annotated data, crisis informatics researchers still cannot fully utilize the capabilities of different computational methods. To overcome these issues, we present to research community a corpora consisting of labeled and unlabeled crisis-related Twitter messages. Moreover, we also provide normalized lexical resources useful for linguistic analysis of Twitter messages.

\section{Conclusions}
We present Twitter corpora consisting of over 52 million crisis-related tweets collected during 19 crisis events.
We provide two sets of annotations related to topic-categorization of the tweets and tagging out-of-vocabulary words and their normalizations. We build machine-learning classifiers to empirically validate the effectiveness of the annotated datasets. We also provide word2vec word embeddings trained on 52 million messages. We believe that these resources and the tools built using them will help improve automatic natural language processing of crisis-related messages and eventually be useful for humanitarian organizations.
\section{References}
\bibliographystyle{lrec2016}
\bibliography{references}

\balance
\end{document}